\documentclass[a4paper,11pt]{article}

\usepackage[english]{babel}
\usepackage[utf8x]{inputenc}
\usepackage{amsmath}
\usepackage{amsfonts}
\usepackage{graphicx}
\usepackage[colorinlistoftodos]{todonotes}
\usepackage[colorlinks=true, allcolors=blue]{hyperref}
\usepackage[font=footnotesize,labelfont=bf]{caption}
\usepackage[font+=footnotesize]{subcaption}

\usepackage{url}       
\usepackage{times}
\usepackage{natbib}
\usepackage[margin=25mm]{geometry}
\usepackage{multirow}
\usepackage{enumitem}
\setlist{nosep}

\usepackage{covington}

\newsavebox{\largestimage}

\definecolor{darkblue}{RGB}{0, 0, 139}
\definecolor{blue}{RGB}{0, 0, 255}
\definecolor{lightblue}{RGB}{173, 216, 230}
\definecolor{olivedrab3}{RGB}{154, 205, 50}
\definecolor{lightgreen}{RGB}{144, 238, 144} 

\title{Towards an Inferential Lexicon \\ of Event Selecting Predicates for French}

\date{}

\author{Ingrid Falk and Fabienne Martin\\
       Universität Stuttgart\\
       \small{\texttt{first.second@ling.uni-stuttgart.de}}
}

\begin{document}
\maketitle
\thispagestyle{empty}
\pagestyle{empty}

\begin{abstract}
  \noindent We present a manually constructed seed lexicon encoding the
  \textit{inferential profiles} of  French event selecting predicates across different uses. The inferential profile
  \citep{karttunen71:_implic_verbs} of a verb is designed to capture
  the  inferences triggered by the use of this verb in context.  It
  reflects the influence of the clause-embedding verb on the \textit{factuality}
  of the event described by the embedded clause.  The 
  resource developed provides evidence for the following three hypotheses: (i) \textit{French implicative verbs have an aspect dependent profile} (their inferential profile varies with outer aspect), while \textit{factive verbs have an aspect independent profile} (they keep the same inferential profile with both imperfective and perfective aspect); (ii)  \textit{implicativity decreases with imperfective aspect}: the inferences triggered by French implicative verbs combined with perfective aspect are often weakened when the same verbs are combined with imperfective aspect; (iii) \textit{implicativity decreases with an animate (deep) subject}:  the inferences triggered by a verb which is implicative with an inanimate subject are weakened when the same verb is used with an animate subject.
The resource additionally shows that verbs with different inferential
  profiles display clearly distinct sub-categorisation patterns. In particular,  verbs that have both factive and implicative readings are shown to prefer infinitival clauses in their implicative reading, and tensed clauses in their factive reading.
\end{abstract}

\thispagestyle{empty}
\pagestyle{empty}

\section{Introduction}
\label{sec:introduction}

Texts not only describe events, but also encode information conveying whether the events described correspond to real situations in the world, or to uncertain, (im)probable or (im)possible situations. This level of information concerns \textit{event factuality}. 
The factuality of an event expressed in a clause results from a complex interaction of many different linguistic aspects.  It depends, among others, on the explicit polarity and modality markers, as well as on the syntactic and semantic properties of other 
expressions involved (among them verbal predicates). In this study,  we are concerned with one of these parameters, namely the predicates selecting event-denoting arguments (e.g. \textit{manage to P}), which  contribute in a crucial way to \textit{lexically} specify event factuality.  \cite{sauri2012}  call these verbs `event selecting predicates', or ESPs.  For example, in ``Kim \textit{failed to} \underline{reschedule} the meeting'', the ESP \textit{fail to} turns the embedded \underline{reschedule}-event into a counter-fact. 

In previous work, \cite{sauri2012} encoded into a lexical resource the  \textit{inferential profile} of English ESPs, that is, their influence on the factuality of the embedded events, and showed that this resource could successfully be used to automatically assess the factuality of events in English newspaper texts. 

Our long-term goal is the automatic detection of event factuality in French texts. 
Given that \citet{sauri2012}'s representation of event modality and their automatic factuality detection method is  language independent, it can also be used for French event factuality detection. We plan to use this approach and describe here our efforts to bootstrap the required lexical ESP resource for French. 

The lexical resource built by \citet{sauri2012} for English is not public, and therefore cannot be used as a starting point for a similar French lexicon. Additionally, Romance languages like French raise a further issue, for, as will be shown in Section~\ref{sec:building-esp-lexicon}, outer (grammatical) aspect interferes with the inferential profile of ESPs. To achieve our long-term goal, we therefore firstly need to build a lexical resource  providing an inferential profile for French ESPs, starting from a seed set of suitable verbs. This resource should specify the inferences a predicate triggers about the event described in the embedded clause (henceforth \textit{embedded event}). For example, it should specify that the perfective form of \textit{échouer à P} `fail to P'  triggers the inference that the embedded event is a counter-fact under positive polarity, whereas the embedded event is entailed when the predicate is used under negative polarity. 

One of the challenges raised by such a lexicon concerns the polysemy of ESPs and the fact that their inferential profile is likely to vary with each use and/or syntactic frame. For instance, \textit{Peter didn't remember \textbf{to} P } and \textit{Peter didn't remember \textbf{that} P} trigger very different inferences about the embedded event (a counter-fact in the former case, and a fact in the latter).  In order to address this challenge,  we collected each  use for these verbs as they are delineated in available detailed syntactic-semantic valency lexicons for French, and calculated an inferential profile for each of them.\footnote{\label{fn:reading}In this paper, we call \textit{readings} of a verb the different verb-valence pairs \textit{and/or} different senses  delineated for a same lemma in these lexicons.}   
The  additional advantage of this method is that we can make use of the detailed syntactic-semantic features encoded for each use in these lexicons. Also, it led to the interesting observation that verbs whose inferential profile varies with the reading selected (including its argument structure) are very pervasive among French ESPs, which confirms the need to   distinguish  between particular senses and/or syntactic frame combinations
an ESP may instantiate. It also revealed interesting correlations between inferential profiles on one hand, and particular sets of  syntactic/semantic properties on the other.

The paper is structured as follows. We first introduce in Section~\ref{sec:related-work}  the two strands of research on which our work is based. We then describe our data and experiments in Section~\ref{sec:building-esp-lexicon} and discuss the resulting findings in Section~\ref{sec:results}.

\section{Related Work}
\label{sec:related-work}

We rely on two important bodies of research. The first is centred around FactBank \citep{sauri2009factbank,sauri2012}, a corpus of English newspaper texts annotated with information concerning the factuality of events. The second is a long standing   research project on English predicates with sentential complements led  at Stanford University \citep{karttunen1971logic,karttunen71:_implic_verbs,nairn2006computing,karttunen2012simple,karttunen2016learning}. We briefly introduce these lines of research in the following subsections.

\subsection{FactBank}
\label{sec:factbank}

The English FactBank is built on top of the English TimeBank corpus \citep{pustejovsky2005temporal} by adding a level of semantic information. The factuality information encoded in TimeBank and relevant for our work is the ESPs projecting a
factual value on the embedded event by means of subordination links (or \textsc{slink}s).
In TimeBank, a total of 9~488 events across 208 newspaper texts have been manually identified and annotated. FactBank assigns additional factuality information to these events.
More specifically, it indicates for each event (i) whether its factuality is assessed by a source different from the text author (which is the case with e.g. \textit{confirm P}, but not with \textit{manage to P}) and (ii)  the degree of factuality the new source and the text author attribute to the event (for instance, \textit{Peter affirmed P} presents \textit{P} as certain for Peter, but does not commit the text author to \textit{P} in a specific way).
\citet{sauri2009factbank} distinguish six `committed' factuality values (i.e. values to which a source is committed) and one `uncommitted' value, which are shown in Table~\ref{tab:NSIPlex}.
\begin{table*}[ht]
    \centering
    \scalebox{0.8}{
    \begin{tabular}{|cc|ccc|ccc|ccc|ccc}
      \hline
      \multicolumn{14}{|c|}{Contextual factuality} \\ \hline
      & & \multicolumn{3}{c|}{CT}&\multicolumn{3}{c|}{PR}&\multicolumn{3}{c|}{PS}&\multicolumn{3}{c|}{U}\\\hline
      \multicolumn{2}{|c|}{polarity} & $+$ & $-$ & u & $+$ & $-$ & u & $+$ & $-$ & u & $+$ & $-$ & \multicolumn{1}{c|}{u} \\
      \hline\hline
      \multicolumn{1}{|c}{\textit{manage}} & & CT$+$ &  CT$-$ &  CTu & PR$+$ & PR$-$ & PRu & PS$+$ & PS$-$ & PSu & Uu & Uu & \multicolumn{1}{c|}{Uu} \\
      \hline
      \multicolumn{1}{|c}{\textit{fail}} & & CT$-$ &  CT$+$ &  CTu & PR$-$ & PR$+$ & PRu & PS$-$ & PS$+$ & PSu & Uu & Uu & \multicolumn{1}{c|}{Uu} \\
    \hline
    \end{tabular}
    }
\caption{Sample lexical entries for NSIPs. CT, PR and PS  signify certain, probable and possible respectively, U (and/or u) unspecified (unknown or uncommitted)}\label{tab:NSIPlex}
\end{table*}
\citet{sauri2012}  present an algorithm, called \textit{DeFacto}, which
assigns to each TimeBank event  a factuality profile consisting of (i)
its factuality value, (ii)
the source(s) assigning the factuality value to that event
and (iii)  the time at which the factuality value assignment takes place. 
The algorithm assumes that events and relevant sources are already identified and computes the factuality profile of events by 
modelling the effect of factuality relations across levels of syntactic embedding.
It crucially relies on three lexical resources which the authors developed manually for English. 
The first is a list of negation particles (adverbs, determiners and pronouns) which determine the \textit{polarity} of the context while
the second resource aims to capture the influence of epistemic \textit{modality} on the event.
The third resource is the most complex one and accounts for the influence on the event factuality value in cases where the event is embedded by  an ESP. Saurí and Pustejovsky distinguish  two kinds of ESPs: \textit{Source Introducing Predicates (SIPs)} introduce a new source in discourse (e.g. \textit{suspect/affirm}); \textit{Non Source Introducing Predicates (NSIPs)} do not 
(e.g. \textit{manage/fail}).
As part of their lexical semantics, SIPs determine (i) the factuality value the new source (the `cogniser')  assigns to the event described by the embedded clause, and (ii) the factuality value assigned by the text author (i.e. the `anchor') to the same event. NSIPs, on the other hand, determine event factuality  wrt. a unique source, the anchor.
In addition, the assessment of event  factuality  wrt. the relevant source(s)  varies with the polarity and modality present  in the  context of the ESP.
Table~\ref{tab:NSIPlex} illustrates the lexicon layout through sample entries for the NSIPs \textit{manage} and \textit{fail}.\footnote{The lexicon layout for SIPs, less relevant for our study, is very similar except that an SIP lexicon entry must also provide factuality values for the cogniser source in addition to the anchor.}
Given the lexical entry for \textit{fail to} shown in Table~\ref{tab:NSIPlex}, the factuality of the \textit{reschedule}$_e$ event in  ``Kim \textit{failed to} \underline{reschedule}$_e$ the meeting'' can be derived as follows. Since in the embedded clause, there are no polarity or modality particles which could influence its factuality, we assume that the contextual factuality of the embedded clause is CT$+$. The corresponding cell in the \textit{fail to} row in Table~\ref{tab:NSIPlex} is CT$-$, i.e. the event is counter-factual.


\subsection{The lexicon resource from  the \textit{Language and Natural Reasoning} group (Stanford)}
\label{sec:fact-impl-verbs}

Another strand of research is represented by 
\citet{nairn2006computing}. Those authors developed a semantic classification of English event selecting predicates, according to their effect on the factuality of their embedded clauses when used in positive or negative polarity. All those verbs are factive or implicative verbs, and therefore non source introducing predicates.\footnote{Factives are predicates that trigger the same entailment under both positive and negative polarity. Implicatives are non-factive predicates that trigger an entailment  under at least one polarity.} In many cases, a single lemma is assigned different entries varying with the syntactic type of the embedded clause (e.g. \textit{remember that/remember to} are assigned two different entries).  This classification is shown in  
Table~\ref{tab:factimp}.
\begin{table}
  \centering
  \scalebox{0.8}{
  \begin{tabular}{|c|cc|r@{$|$}lr@{$|$}l|c|}
\hline
  & \multicolumn{2}{c|}{Polarity of \textsc{esp}} & \multicolumn{4}{c|}{Signatures} & Sample \\
  & $+$ & $-$ & \multicolumn{4}{c|}{} & predicate \\ \hline\hline
  2-way & $+$ & $-$ & $++$&$--$ & $1$&$-1$ & \textit{manage to} \\
  implicatives & $-$ & $+$ & $+-$&$-+$ & $-1$&$1$ & \textit{fail to} \\
  \hline
  1-way & $+$ & $n$ & $++$&$-n$ & $1$&$n$ & \textit{force to} \\
  +implicatives & $-$ & $n$ & $+-$&$-n$ & $-1$&$n$ & \textit{refuse to} \\
  \hline
  1-way & $n$ & $-$ & $+n$&$--$ & $n$&$-1$ & \textit{attempt to} \\
  -implicatives & $n$ & $+$ & $+n$&$-+$ & $n$&$1$ & \textit{hesitate to} \\
  \hline
  factives & $+$ & $+$ & $++$&$-+$ & $1$&$1$ & \textit{forget that} \\
  counter-factives & $-$ & $-$ & $+-$&$--$ & $-1$&$-1$ & \textit{pretend that} \\ \hline
  Neutral & $n$ & $n$ & $+n$&$-n$ & $n$&$n$ & \textit{want to} \\       \hline
  \end{tabular}
  }
  \caption{Semantic classification of clause-embedding verbs wrt. the effect of the polarity of the main clause (ESP, head row) on the factuality of the embedded clause (embedded event, subsequent rows). $n$ stands for \texttt{none}.}
  \label{tab:factimp}
  \vspace*{-8ex}
\end{table}
We illustrate how the table works through concrete examples.

In example 
(\ref{fail:pos}), 
the ESP \textit{fail to} has positive polarity. We obtain the factuality of the embedded event (\textit{reschedule}) by retrieving from the polarity $+$ column in Table~\ref{tab:factimp} the polarity value in the \textit{fail to} row, which is `$-$', i.e. the meeting is not rescheduled (has factuality CT$-$). 
For (\ref{fail:neg}), 
the factuality must be retrieved from the polarity $-$ column resulting in `+', i.e. a factuality of CT$+$ (the meeting is rescheduled). 
\begin{examples}
\item \label{fail:pos} Kim \textit{failed to} \underline{reschedule} the meeting.
\item \label{fail:neg} Kim did not \textit{fail to} \underline{reschedule} the meeting.
\end{examples}
The effect of a predicate on the factuality of its embedded clause is  represented more concisely through a ``signature''. For instance, the  signature of  factive verbs as \textit{forget that} is  `$++|-+$' and even more concisely `$1|1$' (Read: `if positive polarity, event happens; if negative polarity, event happens'). 

\citet{nairn2006computing} compiled a list of roughly 250 English verbs  found to carry some kind of implication: a positive or negative entailment, a factive or a counter-factive presupposition%
\footnote{These resources are available at \url{https://web.stanford.edu/group/csli_lnr/Lexical_Resources/}.}.  The resource makes a difference between entailments (marked by `$+$' or `$-$'), and strong inferences (marked by  `$+^{*}$' or `$-^{*}$'). More recently, \cite{karttunen2016learning} refined the annotation further by using \textit{probabilistic} signatures, allowing to reflect the strength of the inference in a more fine-grained way.
Thus, for example, the predicate \textit{be able} is assigned the signature $0.9|-1$, in order to capture the fact  that under positive polarity, it triggers a strong  (but defeasible) inference rather than an entailment.\footnote{Probabilistic signatures provided in \citet{karttunen2016learning} are rough counts based on the inspection of a lot of examples found on the Internet and the COCA corpus (L. Karttunen, p.c.).}
In Saurí and Pustejovsky's terms, probabilistic signatures convey less than certain factuality values.\footnote{However, while in Saurí and Pustejovsky's annotation, uncertainty is expressed through the two discrete values (PR and PS), probabilistic signatures represent uncertainty on a continuous scale between 0 and 1.}

The signatures assigned in the lexical resource from the Stanford group are calculated for the verb combined with the simple past. This choice seems implicitly  justified by the fact that for many ESPs, no alternative past tense form is available. For instance, it has been observed for  implicative verbs that their progressive form is odd, see e.g.  \textit{?John was managing to eat the pizza} \citep[cf.][]{Bhatt1999}. Also, many factive verbs are stative, and therefore equally odd with the progressive (\textit{*John was knowing the answer}).\\
\indent For  our experiments,  we decided to start with the French counterparts of the English verbs in this resource, assuming that the inferential profile may be roughly extensible to French verbs under their perfective form. We then looked at the sentences in the French TimeBank using these French  ESPs. We first briefly introduce the French TimeBank  before describing our data, experiments and  findings.



\subsection{The French TimeBank}
\label{sec:french-timebank}

The French TimeBank \citep{bittar2010building,bittar2011french} is built on the same principles as the English TimeBank, but introduces additional markup language to deal with linguistic phenomena not yet covered and specific to French. Most relevant to this study 
is the fact that  
 most French  ESPs can be fully inflected and fall within the scope of aspectual  operators --- most French  modal auxiliaries, but also most implicative and factive verbs are fully acceptable in their perfective and imperfective forms. For instance, the French counterpart of \textit{manage to}, namely \textit{réussir à}, accepts both the perfective and the imperfective (see e.g. the example (\ref{reussir:imp}) below). Probably for this reason, all these verbs are also marked up as events in the French TimeBank; in particular, modal verbs are marked up as events of the `modal' subclass (see \cite{bittar2011french}, while in the English TimeBank, modal predicates are not marked up as events.
Lastly, the TimeML schema was adapted to represent the grammatical tense/aspect system of French, and to account 
e.g. for the \textit{imparfait} (the imperfective aspectual morphology), not grammaticalised in English. Since  TimeBanks mark up events and (temporal) relations between them, the French TimeBank offers a sample of ESPs used in French newspaper texts, together with some typical embedded events. \\
\indent Note that the French TimeBank as well as FactBank not only mark up events realised by VPs, but also events realised by NPs. For our experiments, we also annotated the way the verbs under investigation  influence the factuality of events described by one of their event-denoting NP arguments. For instance, \textit{Cela a garanti la survie des passagers} `this ensured the survival of the passengers' and \textit{Cela a garanti que les passagers survivent} `This ensured that the passengers survived' both entail that  the passengers survived, and the annotation aims to capture  parallelisms of this type.


\section{Towards an  ESP lexicon}
\label{sec:building-esp-lexicon}

We started with the observation that  the inferential classification developed by \cite{nairn2006computing} and described in Section~\ref{sec:fact-impl-verbs} can be used to bootstrap an English ESP lexicon.
Based on the signature of a predicate and its polarity in a given sentence, we can determine the factuality of the embedded event in that sentence. The classification in Table~\ref{tab:factimp} can be  straightforwardly ``plugged'' into the ESP lexical resources  illustrated in Table~\ref{tab:NSIPlex}:
For a given ESP for which a lexical entry has to be set up (\textit{eg.} \textit{fail to}), the factuality value conveyed on the embedded event can be retrieved from Table~\ref{tab:factimp} whenever the corresponding table entry is not $n$ (neutral).
In case it is, this shows that the ESP has no effect on the factuality of the embedded event; its polarity value therefore remains unspecified.

\subsection{Our Data}
\label{sec:data}

To build our seed lexicon, we started from the ESPs marked up in the French TimeBank. From these predicates, we selected those 49 verbs which occurred as translations of the English ESPs derived from the inferential classification of \citet{nairn2006computing}. This way, we could use the inferential information from the English classification, and compare it to the French counterparts of these English ESPs.
We assigned  inferential signatures to each reading  of these French ESPs, and this
with two research questions in mind:
\begin{itemize}
\item [i.]Does the inferential behaviour (the signature) vary with the animacy of the external argument and/or with outer aspect?
\item [ii.]Can we differentiate the main sub-types of inferential signatures (factive, implicative, etc.) by  specific subsets of semantic/syntactic properties?
\end{itemize}
\noindent \textbf{Lexical entries.} For the selected 49 verb types we extracted all readings from two French lexicons: ``Les Verbes Français'', henceforth \textsc{lvf} \citep{LVF2007}, and an electronic version of the Lexicon-Grammar tables, henceforth \textsc{lglex} \citep{tolone:tel-00640624,constant:hal-00483662}. We chose these resources because they provide detailed mor\-pho\--syntactic and
semantic descriptions for each reading  of a verb rather than giving an (ambiguous) verb type based description. 
The \textsc{lvf} is based on a traditional French lexicon where the different lexical entries (called here readings, see fn \ref{fn:reading}) for a verb type have been assigned additional detailed and systematic morpho-syntactic, semantic and valency information.
In \textsc{lglex}, each verb type is associated with one or more tables which represent its meaning and behaviour with respect to valency. We consider each such lemma-table association to represent a particular reading of the verb. 
For our seed lexicon, we extracted lexical entries  from the \textsc{lvf} and lemma-table pairs from the \textsc{lglex}. We merged duplicate entries from the two lexicons.  The remaining entries were aligned  such that each entry in our lexicon represents a different reading. 
\noindent \textbf{Annotation.} We obtained $\approx 930$ readings, which we manually filtered, keeping only the ESP readings. The remaining $\approx 170$ readings were manually assigned probabilistic signatures (gathered in a file available on line) by an expert (the second author of this paper).\footnote{At  \url{https://docs.google.com/spreadsheets/d/1sgDxfYSh9lN0zI2hwfq2bG6TdA7EvusaHl5irmWC2QE/edit?usp=sharing}} The annotation was done on the basis of the sentences by which the \textsc{lvf} exemplifies each delineated use of a verb, and on the inspection of natural occurrences of the relevant form in the internet and corpora (such as Frantext).\footnote{The exemplifying sentences  from the \textsc{lvf} can also be found online at \url{http://rali.iro.umontreal.ca/rali/?q=en/node/1238}} When the exemplifying sentence contains elements that can affect the event factuality independently from the ESP itself (modal verbs,  NPs biased towards a generic interpretation, etc.), these elements were abstracted away for the annotation.   We chose probabilistic signatures, since we aimed  to distinguish between entailments, graded inferences (strong vs. weak) and the absence of inference (neutral). We distinguished 4 intermediate positive values, namely 0.9 (very likely/almost certain), 0.8 (likely), 0.7 (very possible/almost likely) and 0.6 (quite possible), and 4 intermediate negative values (-0.9 for very unlikely, etc.). However, for the experiments, we only considered 0.9 to indicate a `strong inference', and the values 0.6 to 0.8 were kept undistinguished from the neutral case (absence of inference).\footnote{In the future, we plan to develop guidelines and explore automatic methods that could assist and support the manual annotations. However,  our manual annotations already sufficed to test interesting hypotheses about  the impact of outer aspect, animacy and syntax on the inferential profile of ESPs, as we will see in the next section.}


For Romance languages like French,  it is known that inferential profiles vary with outer aspect \citep{hacquard2006aspects}. Also, cross-linguistic data tend to suggest that   the inferential profile also varies with the animacy of the (deep) subject \citep{martin2012modality}.\footnote{The sentences tested were mostly active, but some of them were passive. It is the animacy of the external argument (deep subject) that has been shown to matter for the inferential profile of ESPs.} We therefore distinguished three cases:
(i) perfective aspect, animate subject ; (ii) perfective aspect, inanimate subject; (iii) imperfective aspect. Distinguishing again  between two types of subject for the imperfective  was not justified, since most readings (around 80\%) were only compatible with a single type of subject to begin with (either animate, or inanimate, but not both).\footnote{131 out of 168 annotated readings were acceptable with either an animate or an inanimate subject (but not both) in the perfective. Only 27  (16\%) of these 168 readings  which were assigned a signature with the imperfective were acceptable with both an animate and an inanimate subject.} This  reflects the fact that the \textsc{lvf} and  the \textsc{lglex} often differentiate readings according to the (non)-animacy of the subject. This means that in practice, we typically assigned two signatures  instead of three for each reading, namely one for the perfective, either with an animate subject, or with an inanimate one, and one for the imperfective. The third possible case was typically not applicable. So for instance, when a reading compulsorily selects for an animate subject, we assigned one signature for the `animate perfective' context, and assigned none for  the `inanimate perfective' context (introducing the value NA, for `not applicable'). In this case, if a signature is assigned for the  `imperfective' context, we know that this imperfective signature is calculated with an animate subject, too. \\ 
\indent In the rare case where both types of subjects were possible for a single reading (16\% of all readings),
we tested the imperfective verb with the subject type that triggers the strongest entailment with the perfective, namely the inanimate subject. Typically, the effect of the subject type on the inference is neutralised with the imperfective.\footnote{See the examples (7)-(9) below for an illustration.}\\
\indent Table~\ref{tab:obliger02} shows the assigned signatures for the reading \textit{02} of the verb \textit{obliger} (`oblige, force') (as delineated in the \textsc{lvf}). We see that, under perfective aspect and with an animate subject, this predicate triggers, under positive polarity, a strong albeit defeasible inference (there is a high probability that the embedded event is a fact). Under negative polarity, no inference is triggered. Under perfective aspect  with an inanimate subject, \textit{obliger 02} entails the embedded event under positive polarity, and triggers no inference under negative polarity.
Finally, when used with imperfective aspect, \textit{obliger 02} is inferentially neutral, i.e. it triggers no inference about the embedded event (both with animate and inanimate subjects).
\begin{table}
  \centering
  \begin{tabular}{|c|c|c|}
\hline
    \multicolumn{3}{|l|}{Pierre/cela a obligé Marie à partir.}\\
    \multicolumn{3}{|l|}{`Peter/something force-\textsc{past-pfv.3sg} Mary to go.'}\\
\hline
\textsf{PFV+anim} & \textsf{PFV-anim} & \textsf{IMP} \\
\hline
$0.9|n$ & $1|n$ & $n|n$ \\
\hline
  \end{tabular}
  \caption{Assigned signatures for reading \textit{obliger 02} (`force/oblige').}
  \label{tab:obliger02}
  \vspace*{-8ex}
\end{table}

\begin{figure}
  \centering
\hspace*{-2cm}
  \includegraphics[width=20cm]{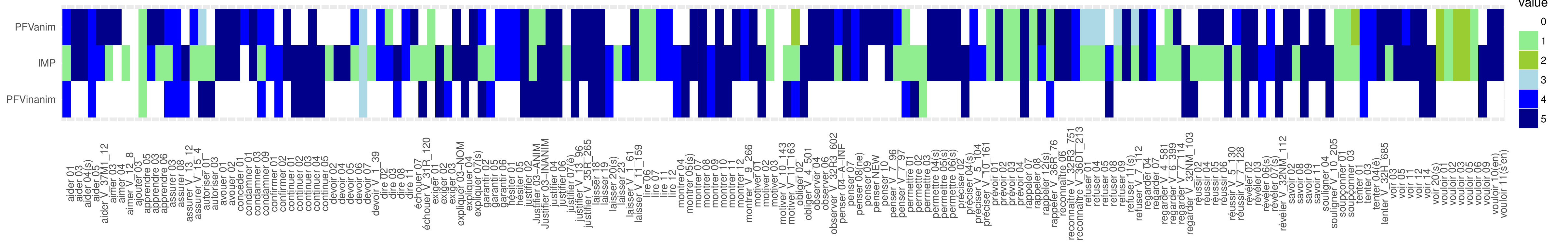}
  \caption{Inferences triggered by French readings when used with perfective aspect and animate or inanimate subject or with imperfective aspect.}
  \label{fig:signatures}
\end{figure}

 Figure~\ref{fig:signatures} gives an overview of the inferences triggered by these 170 readings according to the assigned signatures in the three contexts delineated (\textsf{PFV+anim}: perfective aspect with animate subject; \textsf{IMP}: imperfective aspect with any kind of subject; \textsf{PFV-anim}: perfective aspect with inanimate subject). 
The figure shows a coloured box for each assigned signature. The columns in the matrix represent the  annotated readings  (in alphabetic order)\footnote{In the readings labelled with a simple number (e.g. \textit{aider 01}), the number is the one given in the LVF lexicon, accessible online at \url{http://rali.iro.umontreal.ca/rali/?q=en/node/1238}. When the readings are labelled as in \textit{lemma\_V\_T\_R}, they are extracted from the Lexicon-Grammar tables. \textit{T} designates the table name and \textit{R} the row the verb occurred in.} and the rows correspond to the aspectual and animacy values used when testing the triggered inferences, namely \textsf{PFV+anim}, \textsf{IMP} and \textsf{PFV-anim}. A white box indicates that  the test was not applicable (e.g., the verb is inherently agentive under the relevant reading and is therefore  not annotated with an inanimate subject), or returns agrammaticality (e.g., the verbal reading combined with perfective aspect generates an ungrammatical sentence).
As just mentioned above, a very large number of readings have either a signature for the \textsf{PFV+anim} context, or one for the \textsf{PFV-anim} one, but not for both. In the \textsf{IMP} contexts, almost all readings were assigned a signature, since we did not differentiate between subject types in this context.
Interestingly, more signatures are assigned in the  \textsf{PFV+anim} context than in the  \textsf{PFV-anim} context, which suggests that a substantial number of ESPs tend to be inherently agentive.\\
\indent The colours reflect the ``strength'' of the triggered inference. The strongest inference (value 5) is triggered whenever the signature shows a maximal (deterministic) inference (1 or $-$1) under both polarities (i.e. signatures of the type $x|y$ with $|x|=|y|= 1$). These correspond to the classes of factives, counter-factives and 2-way implicatives displayed in Table~\ref{tab:factimp}. Weaker inferences (value 4) are reflected by signatures where there is a maximal inference (1 or $-$1) under at least one polarity (signatures $x|y$ where $|x| = 1$ or $|y| = 1$, 1-way implicatives in Table~\ref{tab:factimp}). The next value (3) on the inference scale represent (`2-way quasi-implicative') signatures where there is a strong, albeit not maximal and thus defeasible, inference under both polarities ($|x|=|y|=0.9$\footnote{We only distinguish between maximal, strong and neutral inferences, with signature values $|x|=1$, $|x|=0.9$ and $x=0$ respectively. As mentioned before, the annotation has been performed also using intermediate values, which we aim to  use in later work.}). Even weaker inferences (value 2, `1-way quasi-implicative') are those where at least under one polarity the reading is associated with a strong inference ($|x|=0.9$ or $|y|=0.9$). Finally, the absence of inference  (or neutral inference, value 1) is represented by \textit{n}.
Table~\ref{tab:col2inf} sums up the mapping between colours, strength of inference, signatures and inferential classes.
\begin{table}
  \centering
  \begin{tabular}{|c|c|c|r|c|}
\hline
    \textbf{col} & \textbf{description} & \textbf{sample sign.} & \textbf{\#sign.} & \textbf{inferential classes} \\ \hline
\colorbox{darkblue}{\textcolor{white}{5}} & max inference & $1|1$, $1|-1$ & 177 & (counter-)factives, 2-way implicatives \\
\colorbox{blue}{\textcolor{white}{4}} & max under 1 polarity & $1|n$, $0.9|-1$ & 77 & 1-way implicatives \\
\colorbox{lightblue}{3} & strong, not max, under 2 polarities & $0.9|-0.9$ & 9 & 2-way quasi implicatives \\
\colorbox{olivedrab3}{2} & strong, not max, under 1 polarity & $0.9|n$, $n|-0.9$ & 8 & 1-way quasi implicatives \\
\colorbox{lightgreen}{1} & neutral, no inference & $n|n$ & 78 & neutral \\
\hline
0                & not applicable or not grammatical & NA or UNGR & 161 & \\
\hline
  \end{tabular}
  \caption{Overview of mappings between colours, strength of inference, signature and inferential classes.}
  \label{tab:col2inf}
  \vspace*{-8ex}
\end{table}
As shown by the general colour pattern,  inferences under perfective aspect tend to be stronger than those under imperfective (and this even for many implicative verbs), an observation confirmed by a closer look at the data, as we will see in Section~\ref{sec:results}.
\section{Results and discussion}
\label{sec:results}

We first compared the signatures of the French predicates with their rough English counterparts classified by \citet{nairn2006computing}. We found that in many cases, the signatures of the English predicates matched those of their French counterparts under their perfective form. For example, 
the implicative signature `$1|-1$' of \textit{manage to} is inherited by its translation \textit{réussir à} combined with a past perfective morphology.\footnote{Our translation approach  raised additional   issues detailed in \citet{W17-1803}.}
  Second, we aimed to test  three hypotheses  about the semantic and syntactic properties that potentially contribute to drive the inferential profile of the readings:

\begin{enumerate}
\item\label{item:factimp} \textit{implicative verbs have an aspect dependent profile} (their inferential profile varies with outer aspect); \textit{factive verbs have an aspect independent profile} (they keep the same inferential profile with both imperfective and perfective aspect);
\item\label{item:imperf} \textit{implicativity decreases with imperfective aspect}, i.e., the inferences triggered by a verb which is implicative combined with perfective aspect are weakened when the same verb combines with imperfective aspect;
\item\label{item:anim} \textit{implicativity decreases with an animate subject}, i.e., the inferences triggered by a verb which is implicative with an inanimate subject are weakened when the same verb is used with an animate subject.
\end{enumerate}
\noindent Based on counts, we could find evidence in our data supporting  the three hypotheses.\\

\noindent \textbf{Implicatives, but not factives, have an aspect dependent profile.} Factive predicates are  predicates that trigger the same entailment under both positive and negative polarity, i.e. these predicates have either signature $1|1$ or $-1|-1$. Implicatives are  non-factive predicates that trigger an entailment  under at least one polarity (among the potential implicative signatures, we find   e.g. $1|-1$, $1|n$, $-0.7|-1$).
Table~\ref{tab:aspect-factimp} shows that of the 54 readings that were factive with perfective aspect, only 2 had a different signature with imperfective aspect. On the other hand, for  the readings that were implicative with the perfective, almost half of them (36 out of 77 or 48\%)  had a different signature with the imperfective aspect (are only considered here readings for which a signature could be assigned with both perfective and imperfective aspect).
\begin{table}
  \centering
  \begin{tabular}{|c|c|c|}
\hline 
 & with \textsf{IMP} \and \textsf{PFV} signature & \textsf{IMP} sign$\not=$\textsf{PFV} sign \\ 
\hline 
Factives under PFV  & 54 & 2 (4\%) \\ 
\hline 
Implic. under PFV & 77 & 36 (48\%)\\
\hline
  \end{tabular}
  \caption{Influence of outer aspect on factivity.}
  \label{tab:aspect-factimp}
  \vspace*{-8ex}
\end{table}
In other words, outer aspect has virtually \textit{no} influence on factivity. This is reflected in Fig. \ref{fig:signatures} by the fact that the cells with the darkest colour in the \textsf{PFV+anim} row keep this darkest colour in the \textsf{IMP} row. By contrast, the inferential profile of implicative verbs tends to change with aspect, supporting our hypothesis \ref{item:factimp}.

\noindent \textbf{Implicativity decreases with imperfective aspect.} For French, it is standardly assumed that implicative verbs keep their entailment with the perfective (see e.g. \citealt{hacquard2006aspects}). Our data challenge this assumption, and support the alternative hypothesis that implicativity tends to decrease with imperfective aspect.  This hypothesis is illustrated by the data in  examples \ref{reussir:pfv} and \ref{reussir:imp}, with the verb  \textit{réussir} `manage to', that has an implicative profile when combined with perfective aspect.\footnote{The examples are made up for illustration purposes. This use of \textit{réussir} corresponds to the readings \textit{05} and \textit{06} in the \textsc{lvf}.} Recall that implicatives trigger a maximal inference under at least one polarity, i.e. at least under one polarity the signature is $max=1$ or $max=-1$.\footnote{Therefore, an implicative signature $max|x$ or $x|max$ describes a stronger inference than signatures $a|x$ or $x|a$ with $|a|<1$ or $a=n$.}

\begin{examples}
\item\label{reussir:pfv} \gll A ce moment-là, elle {\textbf{a réussi}} à {s'enfuir}. \#Et pourtant, elle {ne s'est pas enfuie} du tout.
at that moment she manage-\textsc{\textbf{pst-pfv}.3sg} to escape and nevertheless she {\textsc{neg} escape-\textsc{\textbf{pst-pfv}.3sg}} at all
\glt `At that moment, she managed to escape. And nevertheless, she didn't escape at all.'
\glend 
\item\label{reussir:imp} \gll A ce moment-là, elle \textbf{réussissait} (encore) à s'enfuir. OK Et pourtant, elle {ne s'est pas enfuie} du tout/ elle {n'essayait} même pas de le faire!
at that moment she manage-\textsc{\textbf{imp}.3sg} (still) to escape {} and nevertheless, she {\textsc{neg} escape-\textsc{\textbf{pst-pfv}.3sg}} at all she {\textsc{neg} try-\textsc{\textbf{pst-imp}.3sg}} even \textsc{neg} to it do
\glt `At that moment, she `was (still) managing' to escape. And nevertheless, she didn't escape at all/ she wasn't even trying to do so.'
\glend 
\end{examples}
In this example, the same predicate is used in two contexts that mainly differ by the aspect morphology on the verb: perfective in (\ref{reussir:pfv}) and imperfective in (\ref{reussir:imp}). However, in (\ref{reussir:pfv}), \textit{réussir} triggers an entailment, whereas the inference triggered in (\ref{reussir:imp}) is defeasible (both through a perfective and an imperfective denial).\footnote{\citet[71]{hacquard2006aspects} claims that the imperfective form of \textit{réussir} is implicative on the basis of the oddity of  sentences like  \textit{Darcy réussissait à soulever cette table, \#mais il ne la soulevait pas.} (`Darcy manage-\textsc{\textbf{pst-imp}.3sg} to lift this table, but wasn't lifting it'). We agree with Hacquard's judgment, but we think that the problem partly comes from the absence of any adverbial providing a reference time combined with the presence of the imperfective in the second clause. A sentence like \textit{A ce moment-là, Darcy réussissait clairement à soulever cette table. Et pourtant, il n'essayait même pas de la soulever} (`At that moment, Darcy clearly manage-\textsc{\textbf{pst-imp}.3sg} to lift to lift this table. And nevertheless, he wasn't even trying to do so.') is completely acceptable.}

The sceptical reader could argue that in (\ref{reussir:imp}), the imperfective sentence is the consequent of an implicit conditional (\textit{A ce moment-là, il réussissait à s'enfuir s'il le voulait} `At that moment, he manage-\textsc{\textbf{pst-imp}.3sg} to flee if he want-{\textsc{\textbf{pst-imp}.3sg}} to').  The absence of entailment would then reflect the fact that this implicit conditional is counter-factual, rather than the influence of aspect on the inferential profile of \textit{réussir}.\footnote{In French, counter-factual conditionals typically have the \textit{conditionnel} in the consequent (and the \textit{imparfait} in the antecedent). However, it is also possible to have an imparfait in both the antecedent and consequent, as also the case in Italian, cf. \citet{Ippolito03}.}
However,  this objection does not apply to the example (\ref{reussir:imp2}) below, where  a conditional antecedent cannot be felicitously added. Additionally, (\ref{reussir:imp2}) contains a past progressive, which, contrary to the imperfective, is not a necessary ingredient of counter-factuality \citep[257]{Iatridou00}. The sentence in (\ref{reussir:imp2})  therefore contains a run-of-the-mill imperfective, and, again, the actuality entailment triggered with the perfective in (\ref{reussir:pfv2}) is lost.    
  
\begin{examples}\label{reussir2}
\item\label{reussir:pfv2} \gll Ana {a réussi} à gagner la partie \#quand {tout à coup}, son rival {l'a faite} échec et mat.
Ana manage-\textsc{\textbf{pst-pfv}.3sg} to win the game when {all of a sudden} her opponent {her make\textsc{\textbf{pst-pfv}.3sg}} chess and mat
\glt `Ana managed to win the game when all of a sudden, her opponent gave her checkmate.' 
\glend 
\item\label{reussir:imp2}  \gll Ana {était en train de réussir} à gagner la partie quand {tout à coup}, son rival {l'a faite} échec et mat.
Ana manage-\textsc{\textbf{pst-prog}.3sg} to win the game when {all of a sudden} her opponent {her make-\textsc{\textbf{pst-pfv}.3sg}} chess and mat
\glt `Ana `was managing' to win the game when all of a sudden, her opponent checkmated her.' 
\glend 
\end{examples}

\noindent As shown in Table~\ref{tab:imp-aspect}, we found strong support in our data for the hypothesis that  the inferential behaviour of \textit{réussir} follows a more general pattern. That is,  in general, the  inferences characteristic of implicative readings  are in fact conditional on the presence of perfective aspect in languages like French, and tend to be weakened or even vanish with imperfective aspect.\footnote{Among the 41 readings that remain implicative with imperfective aspect, most do not express what we call below the actualisation of a causal factor for the truth of \textit{p}, and are therefore not expected to lose their entailment with the imperfective. This is for instance the case of \textit{montrer 04, 09, 11, 12, V-9-266} `show', \textit{penser 07, 09, 09} `think', \textit{vouloir 09} `want'. Also, quite a few of these readings are reflexive (7), a factor that seems to interfere with the inferential profile, too.}
In almost all cases (34 out of 36) where a reading was implicative with a perfective and triggered a different inference with the imperfective, the inference triggered with the imperfective was weaker 
than the one triggered with the perfective.
\begin{table}
 \centering
\begin{tabular}{|c|c|c|}
\hline
 \textsf{IMP}$\leadsto$weaker inference & \textsf{IMP}$\leadsto$stronger inference & no change \\ 
\hline 
44.2\% (34) & 2.6\% (2) & 53.2\% (41)\\ 
\hline 
\end{tabular}
\caption{Inferential behaviour of the 77  readings which are implicative with a perfective. For 34 of them (44.2\%), the inference triggered with perfective aspect is stronger than with imperfective aspect.}
\label{tab:imp-aspect}
\end{table}
Table~\ref{tab:permettre} shows some more examples (from the \textsc{lvf}) where the inferential profile varies with the aspect used. Also, the examples below for the verb \textit{refuser} `to refuse'  illustrate the general fact that even when the perfective verb triggers an entailment with an inanimate subject \textit{only}, this entailment is lost in the \textsf{IMP} even when combined with such an inanimate subject.\footnote{The use of \textit{refuser} with animate subject illustrated in (\ref{refuser}) corresponds to the   \textsc{lvf} reading \textit{09}, and its use with inanimate subject illustrated in (\ref{refuser-2}) corresponds to the  \textsc{lvf}  reading \textit{08}.}

\begin{examples}\label{refuser}
\item \gll Pierre {a refusé}/refusait que {j'ouvre} le tiroir, OK mais je {l'ai ouvert} {quand même.}
Pierre refuse-\textsc{\textbf{pst-pfv}/\textbf{pst-imp}.3sg} that {I open} the drawer {} but I {it open-\textsc{\textbf{pst-pfv}.1sg}} nevertheless
\glt `Pierre refused to let me open the drawer, but I opened it nevertheless.'
\glend                    
\item \label{refuser-2} \gll Le tiroir {a refusé} de {s'ouvrir}, \#mais finalement, il {s'est ouvert} {quand même}.
the drawer refuse-\textsc{\textbf{pst-pfv}.3sg} to {\textsc{refl} open} but {at the end} it {\textsc{refl} open-\textsc{\textbf{pst-pfv}.3sg}} nevertheless   
\glt `The drawer refused to open, but at the end, it opened nevertheless.'
\glend
\item \gll Le tiroir refusait de {s'ouvrir}, OK mais finalement, il {s'est ouvert} {quand même}/ en forçant un peu Ana {l'ouvrait} sans problème. 
the drawer refuse-\textsc{\textbf{pst-imp}.3sg} to {\textsc{refl} open} {} but {at the end} it  {\textsc{refl} open-\textsc{\textbf{pst-pfv}.3sg}} nevertheless in forcing a bit Ana {it open-\textsc{\textbf{pst-imp}.3sg}} without problem
\glt `The drawer `was refusing' to open, but at the end, it opened nevertheless/by forcing a bit Ana was opening it without difficulty.'
\glend
\end{examples}

\begin{table}
  \centering
  \scalebox{0.9}{
  \begin{tabular}{|l|l|r@{$|$}lr@{$|$}l|}
\hline
  \multirow{2}{*}{\textbf{reading}} & \multirow{2}{*}{\textbf{translation}} & \multicolumn{4}{c|}{\textsc{\textbf{aspect used}}}  \\
  && \multicolumn{2}{c}{\textsf{PFV}} & \multicolumn{2}{c|}{\textsf{IMP}}  \\ \hline\hline
  \textit{assurer 03 (la victoire)} & `ensure (the victory)' & $1$&$n$ & $n$&$n$  \\
  \hline
    \textit{échouer 07 (à persuader} \textit{x)'} & `fail (to persuade \textit{x})' & $-1$&$1$ & $n$&$n$  \\
  \hline
      \textit{motiver 03 (x à venir)} & `motivate (\textit{x} to come)' &$0.7$&$-0.7$ & $n$&$n$  \\
  \hline
         \textit{penser 04 (à divorcer)} & `think (about divorcing)' & $-0.7$&$-1$ & $n$&$n$  \\
  \hline
  \end{tabular}
  }
  \caption{Examples of verbs whose inferential profile varies with the outer aspect used.}
  \label{tab:permettre}
  \vspace*{-8ex}
\end{table}

\noindent Why do  implicative verbs (contrary to factive verbs) lose their entailment when combined with \textsf{IMP}?  Recent analyses of  implicative verbs  by \cite{BagliniFrancez16} and \cite{Nadathur16} can help to explain this observation.  According to  Baglini \& Francez' analysis, a \textit{manage p} statement \textit{presupposes} familiarity with a causally necessary but insufficient condition \textit{A} for the truth of \textit{p}, and  \textit{asserts} that \textit{A}  actually caused the truth of \textit{p}. Nadathur extends a modified version of this analysis to the whole class of implicative verbs. The important point for us is that under these analyses,  implicative verbs have an at-issue component (and on this point,  they drastically differ from the traditional analysis held e.g. by \citet{karttunen71:_implic_verbs}, according to which implicative verbs like \textit{manage to P} make no truth-conditional contribution beyond that of their embedded clause). Under these new analyses,  implicative verbs assert what we propose to call an `actualisation event', namely the obtaining/actualisation of the causal factor  \textit{A} for the truth of \textit{p}. Given the `imperfective paradox' \citep{Dowty77}, the imperfective form of such verbs unsurprisingly \textit{suspends} the actualisation event -- similarly to what happens with the imperfective form of overtly causative verbs (e.g. \textit{Trump was causing a new catastrophe when Pence stopped him} does not entail the occurrence of a new catastrophe).  \\
\indent On the other hand, factive verbs like \textit{savoir que p} `know that \textit{p}'  do not assert the obtaining of a causal factor for  the truth of \textit{p}, but rather a mental state having \textit{p} as its object. We therefore do not expect aspect to interfere with their inferential profile.

\noindent \textbf{Implicativity decreases with an animate subject.} 
We showed through the example in Table~\ref{tab:obliger02} (\textit{Pierre/\-ce\-la a obligé Marie à partir}/`Peter/something force-PFV-3SG Mary to go') that the inference triggered by the perfective form of \textit{obliger 02} (`oblige/force') when used with an animate subject (signature $0.9|n$) is weaker than with an inanimate subject (signature $1|n$). This example  also instantiates a more general pattern  observed in our data. Firstly, if only coloured boxes in Fig. \ref{fig:signatures} are considered, the  `\textsf{PFV-anim}' row contains altogether darker colours (representing stronger inferences) than the `\textsf{PFV+anim}'. This is reflected in the respective average inference value for these two contexts: 4,53 for \textsf{PFV-anim} v. 3,92 for \textsf{PFV+anim}. Secondly, the same pattern is confirmed when we look at particular verbs under their different readings.
Among the 49 verbs in our inferential lexicon, 13 received a signature  in both \textsf{PFV+anim}  and \textsf{PFV-anim}  contexts.
For 8 of these 13 verbs, the readings with inanimate subjects were found to trigger stronger inferences than those with an animate subject. 


Further evidence supporting this hypothesis is related to source introducing predicates (SIPs, defined in Section~\ref{sec:factbank}). SIPs  typically trigger no or only weak inferences on the (non-)occurrence of the embedded event in the world of evaluation (for instance, \textit{Peter believed that P} does not say much  about whether \textit{P} is a fact in the actual world --- it mainly indicates that Peter believes \textit{P} to be sure). Of the 13 predicates with SIP readings, 10 also had readings with an inanimate subject.
Unsurprisingly, all of these 10 inanimate readings were not SIP readings anymore (since the inanimate subject cannot be a cogniser), but more interestingly, they have then an implicative or quasi-implicative reading. For example, in \textit{Pierre a garanti le succès de l'affaire} `Peter promised  success for the business', \textit{garantir 06} (with animate subject) introduces a new source and therefore tells us nothing about the embedded event in the actual world. On the other hand,  in \textit{Cela a garanti notre survie} `This ensured our survival',  \textit{garantir 05} (with inanimate subject) is \textit{not} an SIP reading, and is implicative.\footnote{Examples taken from the \textsc{lvf}.}

Differently from  implicative verbs that often can take either an animate or inanimate subject, factive predicates seem to require an animate subject much more forcefully. Out of the 42 readings with a factive signature with an animate subject, only 9 readings are acceptable with an inanimate subject.  This restriction imposed on the animacy of the subject by factive verbs probably simply reflects the fact that these predicates are inherently attitude verbs.

\noindent \textbf{Syntactic type of embedded clauses.} For English, it has often been pointed out  that factives and implicatives show different sub-categorisation patterns.  \cite{White2014} mentions that for verbs that have both factive and implicative readings,  the factive reading is often associated with  a tensed (\textit{that-)}clause  (as e.g. \textit{remember that}) whereas the implicative reading is associated with an   infinitival clause (as e.g. \textit{remember to}). For implicative verbs, it has often been observed \citep[by e.g.][]{landau2001elements} that they typically do not take finite (\textit{that-}) clauses (cf. e.g. \textit{*manage/dare that}), but often select infinitival (\textit{to-}) clauses.  
However, to our knowledge, there is no study confirming these correlations by empirical evidence. 

Given that similar correlations are expected to hold for French, we therefore looked at the type of embedded clauses associated with the readings in our French inferential lexicon. More specifically, we checked, for each reading, whether it can sub-categorise the infinitival and tensed clauses listed in Table~\ref{tab:compclauses}.
\begin{table}
  \centering
  \scalebox{0.9}{
  \begin{tabular}{|c|c|c|c|}
\hline
    & & \textbf{example} & \textbf{translation} \\
\hline
   \multirow{3}{*}{infinitival} & \textbf{aInf} & Il \textit{autorise} Pierre \textbf{à sortir}.& `He allows Pierre to go out.'\\
               & \textbf{deInf} & Le tiroir \textit{refuse} \textbf{de s'ouvrir}.& `The drawer refuses to open.'\\
               & \textbf{inf}   & Il \textit{regarde} la pluie \textbf{tomber}.& `He watches the rain fall.'\\
\hline
   finite      & \textbf{que}   & Il \textit{pense} \textbf{que Pierre est sincère}.& 'He thinks Peter is sincere.'\\
\hline
  \end{tabular}
}
  \caption{Types of infinitival and finite embedded clauses for French observed for our data (sample phrases taken from \textsc{lvf}).}
  \label{tab:compclauses}
\end{table}
We firstly extracted this information from our two French valency lexicons. However, since the data collected this way was not completely reliable, we manually corrected the classification thus obtained.

\begin{figure}
  \centering
\savebox{\largestimage}{\includegraphics[height=.22\textheight]{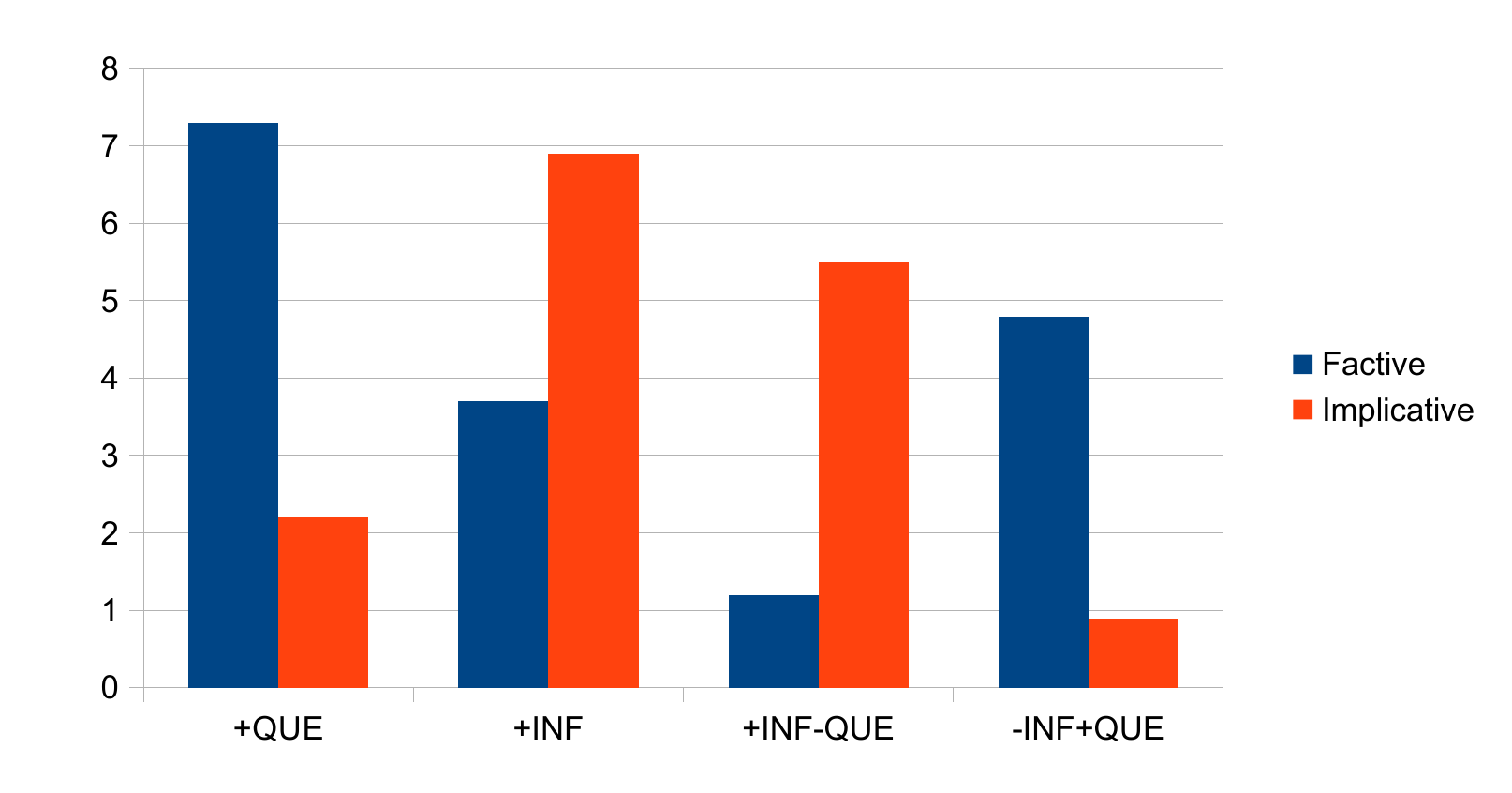}}
  \begin{subtable}[b]{0.48\textwidth}
\centering
\raisebox{\dimexpr.5\ht\largestimage-.5\height}{%
\begin{tabular}{|c|c|c|}
\hline 
& implic. r. (45) & factive r. (41) \\
\hline
+INF & 69\% & 37\% \\ 
+QUE  & 22\% & 73\% \\
+INF $-$QUE & 55\% & 12\% \\
$-$INF +QUE & 9\% & 48\% \\ 
\hline 
\end{tabular}}
  \end{subtable}%
  \begin{subfigure}[b]{0.48\textwidth}
\centering
    \includegraphics[width=\textwidth]{diagramme-slide-EACL.pdf}
  \end{subfigure}
  \caption{Factive and implicative verbs display different syntactic patterns with respect to the types of sub-categorised embedded clauses.}
  \label{fig:que-inf}
  \vspace*{-8ex}
\end{figure}

In our data, 20 verbs had both factive (41) and implicative (45) readings. Figure~\ref{fig:que-inf} shows that there are very clear differences in the types of embedded clauses accepted by implicative vs. factive readings of these `inferentially polysemous verbs': implicative readings tend to be associated with infinitival clauses and to reject tensed clauses, while factive readings show the inverse pattern.  

\section{Conclusion and outlook}
\label{sec:concl-furth-work}

In this work, we developed a seed lexicon associating readings of French clause-embedding verbs with three different probabilistic inferential signatures, varying with outer aspect and the animacy of the subject.
Through these inferential signatures, we concisely represent the inferences that can be drawn from the use of a clause-embedding predicate about the event expressed by the embedded clause given the polarity used.
The experiments performed on the collected data support three hypotheses. Firstly, they show that in  French where most ESPs have perfective and imperfective forms,  implicativity, but not factivity, depends on perfective aspect. Interestingly, this hypothesis seems extendable to other Romance languages like  Spanish, where implicative verbs also seem to lose the actuality entailment obtained with the perfective when they are used imperfectively, see the contrast below:

\begin{examples}\label{reussir3}
\item\label{reussir:pfv3} \gll Ana logr\'{o} ganar la partida \#cuando, {de repente}, su rival le dio jaque mate.
Ana manage-\textsc{\textbf{pst-pfv}.3sg} win the game when {all of a sudden} her opponent her give-\textsc{\textbf{pst-pfv}.3sg} chess  mat
\glt `Ana managed to win the game when all of a sudden, her opponent checkmated her.' 
\glend 
\item\label{reussir:imp3}  \gll Ana {estaba logrando} ganar la partida OK cuando, {de repente}, su rival le dio jaque mate.
Ana manage-\textsc{\textbf{pst-prog}.3sg} win the game {} when {all of a sudden} her opponent her give-\textsc{\textbf{pst-pfv}.3sg} chess  mat
\glt `Ana `was managing' to win the game when all of a sudden, her opponent checkmated her.' 
\glend 
\end{examples}

\noindent Secondly, they show that implicativity, but not factivity, tends to decrease with the animacy of the subject. Thirdly, they indicate that verbs with both a factive and an implicative reading tend to accept infinitival clauses  and to reject tensed clauses under their implicative readings, while they show the inverse pattern under their factive readings.

\section*{Acknowledgments} We are grateful to the anonymous IWCS reviewers for their detailed and constructive feedback, criticisms and suggestions. This work is part of the project B5 of the Collaborative Research Centre 732 hosted by the University of Stuttgart and financed by the \textit{Deutsche Forschungsgemeinschaft}.

\bibliographystyle{chicago}
\bibliography{main}

\end{document}